\documentclass[runningheads]{llncs}
\usepackage{graphicx} 
\usepackage{tabularx}
\usepackage{booktabs}
\usepackage{amsmath}
\usepackage{float}
\usepackage{color,soul}
\usepackage{amssymb}
\usepackage{url}
\usepackage{colortbl}
\usepackage[table]{xcolor}
\newcolumntype{g}[1]{>{\columncolor{gray!#1}}c}

\begin{document}
\title{Latent Graphs for Semi-Supervised Learning on Biomedical Tabular Data}
\titlerunning{Latent Graphs from Tabular Data}
%
\author{Boshko Koloski\inst{1,2}\and
Nada Lavra\v{c} \inst{1} \and
Senja Pollak \inst{1} \and
Bla\v{z} \v{S}krlj \inst{1}
}
\authorrunning{B. Koloski et al.}
%
\institute{Jožef Stefan Institute, Ljubljana, Slovenia 
\\ 
\and
Jo\v{z}ef Stefan International Postgraduate School, Ljubljana, Slovenia
\email{\{boshko.koloski,nada.lavrac,senja.pollak,blaz.skrlj\}@ijs.si}}
\maketitle              
\begin{abstract}
In the domain of semi-supervised learning, the current approaches insufficiently exploit the potential of considering inter-instance relationships among (un)labeled data. In this work, we address this limitation by providing an approach for inferring latent graphs that capture the intrinsic data relationships. By leveraging graph-based representations, our approach facilitates the seamless propagation of information throughout the graph, effectively incorporating global and local knowledge. Through evaluations on biomedical tabular datasets, we compare the capabilities of our approach to other contemporary methods. Our work demonstrates the significance of inter-instance relationship discovery as practical means for constructing robust latent graphs to enhance semi-supervised learning techniques. The experiments show that the proposed methodology outperforms contemporary state-of-the-art methods for (semi-)supervised learning on three biomedical datasets.

\end{abstract}
\section{Introduction}
Machine learning has undergone remarkable advancements in recent years, transforming numerous domains by enabling computers to learn patterns and make predictions from data. In the early stages of this field, there was a strong emphasis on learning from tabular data \cite{quinlan1986induction,breiman2001random}. Pioneering researchers dedicated their efforts to constructing simple yet interpretable models that capitalized on this data type, yielding impressive performance during inference.
The focus on learning from tabular data stemmed from its ubiquity in various domains, where structured information is readily available in the form of rows and columns. The simplicity and comprehensibility of tabular data make it an ideal starting point for machine learning tasks, allowing for effective modeling and decision-making.
These early approaches to machine learning extracted valuable insights and predictions by leveraging the inherent structure and relationships within tabular datasets. The constructed models exhibit remarkable interpretability, enabling human experts to comprehend and reason for the decision-making processes. This interpretability is pivotal in domains where transparent and accountable decision-making is crucial. In real-world machine learning, labeled data is often scarce but unlabeled data is abundant. To enhance predictive performance, several approaches have been proposed to incorporate this unlabeled data into the learning process. Referred to as semi-supervised methods, these approaches combine supervised learning with unsupervised learning techniques to leverage the untapped potential of unlabeled data. By doing so, they aim to improve the overall predictive capabilities of the models while reducing the reliance on labeled data, ultimately addressing the challenges associated with data scarcity and the high cost of annotation.For example, predictive clustering trees (PCTs) \cite{stepinik2020oblique} learn cluster labels as features, which can be used to enrich the feature set of the training data. This can lead to improved predictive performance, especially when there is limited labeled data available. Contemporary approaches in semi-supervised learning focus on projecting the data into lower-dimensional spaces using techniques such as linear learners like SVD~\cite{10.1145/3369390} or autoencoder~\cite{balabka2019semi,ehsan2017infinite} neural network architectures. These methods exploit dimensionality reduction to capture essential patterns and extract informative representations, enabling enhanced learning and generalization capabilities.

In this work, we present a semi-supervised learning approach that transforms the problem of instance classification into node classification. We first construct a latent graph from the data, and then learn a graph neural network on this graph. This approach allows us to leverage the relationships between instances in the data based on inter-instance similarity to improve the classification accuracy.

The rest of the paper is structured as follows: Section \ref{sec:rel_work} presents an overview of the related work, Section \ref{sec:meth} elaborates on our method, Section \ref{sec:experiments} details the experimental setup and Section 5 presents the obtained results. Finally, the paper presents the conclusions and suggestions for future work in Section 6.

\section{Related work}
\label{sec:rel_work}
Semi-supervised learning is concerned with leveraging weakly-labeled or unlabeled data in addition to labeled data. Early approaches concentrated on employing clustering methods such as KMeans \cite{kmenas} and DBSCAN \cite{dbscan} to learn cluster labels and incorporate them into the learning process \cite{levatić2022semisupervised}. Contemporary methods have harnessed latent space projections achieved through dimensionality reduction techniques such as SVD \cite{wold1987principal}, tSNE \cite{maaten2008visualizing}, and UMAP \cite{mcinnes2018umap}. Initially, a linear projection is learned on the entire dataset, followed by applying a learner on the transformed data. Such approaches have demonstrated efficacy across diverse domains\cite{10.1145/3369390}, with notable applications including Latent Semantic Analysis \cite{koloski2020multilingual}. Alternatively, other approaches focus on learning data reconstruction using autoencoders to enhance the learning process \cite{balabka2019semi,ehsan2017infinite}. The encoded latent representation of the input is then used to train a predictive model.
\par Graphs provide a distinctive means of representing data, offering the potential to enhance the predictive capabilities of statistical and neural learners \cite{xu2018powerful,koloski2022knowledge}. Nevertheless, graphs are not always readily accessible in every scenario, prompting researchers to propose diverse approaches to tackle this challenge. Koloski et al. \cite{koloski2023inducing} focused on inducing a graph based on the similarity of given instances and their closest k-neighbours. Bornstein et al. \cite{kazi2022differentiable} proposed learning the graph in a differentiable end-to-end scenario. Learning on the latent graphs can be done with message-passing architectures like Graph Convolutional Neural Networks (GCN) \cite{kipf2016semi}, which are inherently semi-supervised learners. 
To our knowledge, no work has addressed building latent graphs and learning from them on wide tabular data from the biomedical domain, characterized by a small number of instances described by large number of features.

\section{Methodology}
\label{sec:meth}

Following the framework proposed in our previous work \cite{koloski2023inducing}, this section presents the proposed two-step methodology consisting of a latent graph construction step (Section \ref{subsb:lg}), followed by a classification step implemented through a two-layer graph convolutional network (Section \ref{subsb:gcn}).



\subsection{Latent Graph Construction}
\label{subsb:lg}
Given a dataset consisting of \emph{N} instances, the goal is to construct a graph \(\mathcal{G}(\mathcal{V}, \mathcal{E})\), where \(\mathcal{V}\) represents the set of vertices and \(\mathcal{E}\) represents the set of edges. In our case, the set of vertices corresponds to the instances in the data $|\mathcal{V}|$ = \emph{N}. To create the graph's edges, we calculate the cosine distance between the instances $i$ represented with features as $X^{(i)}$ and instance $j$ as $X^{(j)}$:


\begin{equation*}
\mathbf{cos}(X^{i},X^{j}) = \frac{X^{(i)} \cdot X^{(j)}}{\|X^{(i)}\| \|{X^{(j)}}\|}
\end{equation*}

This allows us to capture the similarity between instances and represent it as edge weights in the graph. The cosine distance metric provides a measure of similarity based on the angle between the instance vectors. The values in the adjacency matrix range from $\{-1,1\}$. In this way, we obtain a full graph (similar to the previous work \cite{koloski2023inducing}); we only keep the edges that have a cosine score greater than some threshold $\theta$ greater than 0, i.e., an edge between examples $i$ and $j$ is constructed as follows: 
\begin{equation*}
e_{ij} = \begin{cases}
1 & \text{if } \mathbf{cos}(X^{i},X^{j}) \geq \theta \\
0 & \text{otherwise}
\end{cases}
\end{equation*}

\subsection{Graph Convolutional Network}
\label{subsb:gcn}

We employ a two-layer Graph Convolutional Network (GCN) \cite{kipf2016semi} to exploit the latent graph structure and learn meaningful representations of the instances. Formally, given an adjacency matrix \( A \) and node features matrix \( X \), a GCN performs node representation learning through a sequence of graph convolutional layers. The node representations are updated at each layer by aggregating information from neighboring nodes. This aggregation is achieved by combining the features of each node \( v \) with its neighbors \( \mathcal{N}(v) \), weighted by the graph structure. The computation can be expressed as:

\[
h_v^{(l+1)} = \sigma \left( \sum_{u \in \mathcal{N}(v)} \frac{h_u^{(l)} W^{(l)}}{\sqrt{{|\mathcal{N}(v)| \cdot |\mathcal{N}(u)|}}} \right)
\]

where \( h_v^{(l)} \) denotes the representation of node \( v \) at layer \( l \), \( \sigma \) is the activation function, \( W^{(l)} \) is the learnable weight matrix at layer \( l \), and \( \mathcal{N}(v) \) represents the set of neighboring nodes of \( v \). Finally, a linear classification layer is applied to predict the probabilities \[
\mathbf{y} = \text{softmax}(W^{(2)}\mathbf{h})
\]
\noindent where \(\mathbf{y}\) represents the predicted class probabilities, \(W^{(2)}\) is the weight matrix of the linear layer, and \(\mathbf{h}\) is the flattened output of the last GCN layer. 

We train our graph convolutional network (GCN) model using the Adam optimizer \cite{kingma2014adam} with a learning rate of 0.01 and weight decay of 5e-4. To prevent overfitting and achieve optimal performance, we employ early stopping \cite {prechelt1998early}. The training is stopped if the validation loss does not improve for 10 epochs. The GCN architecture is implemented using the PyTorch library \cite{paszke2017automatic}.

\section{Experimental Setting}
\label{sec:experiments}

\subsection{Data}

In our study, we performed experiments on a collection of biomedical datasets characterized by a wide format, where the number of columns exceeded the number of instances, featuring instance counts ranging from 32 to 801 and the number of features spanning from 661 to 20,531. Across all datasets, the features are numerical. All datasets except $Multi_A$ are unbalanced. Table \ref{tab:table} contains more comprehensive statistics for the used datasets. 

\begin{table}[H]
\caption{Dataset summary. For each dataset, we report the number of instances, features, classes, and the class distribution.}
    \label{tab:table}
    \centering
    \begin{tabular}{lccc|g{10}|g{30}|g{50}|g{70}|g{90}}
    \toprule
    \textbf{Dataset} &  \textbf{Instances} &  \textbf{Features} &  \textbf{Classes} &
     \multicolumn{5}{c}{\textbf{Class Distribution (\%)}}
     \\
   & & & & Class 1 & Class 2 & Class 3 & Class 4 & Class 5 \\
     \\
  \midrule
$Multi_B$ \cite{tissues} & 32 & 5565 & 4 & 34.38 & 28.12 & 21.88 & 15.62 & - \\
$Breast_B$ \cite{tissues} & 49 & 1213 & 4 & 38.78 & 24.49 & 22.45 & 14.29 & - \\
$DLBCL_C$ \cite{tissues} & 58 & 3795 & 4 & 29.31 & 27.59 & 22.41 & 20.69 & - \\
$Breast_A$ \cite{tissues} & 98 & 1213 & 3 & 52.04 & 36.73 & 11.22 & - & - \\
$Multi_A$ \cite{tissues} & 103 & 5565 & 4 & 27.18 & 25.24 & 25.24 & 22.33 & - \\
$DLBCL_D$ \cite{tissues} & 129 & 3795 & 4 & 37.98 & 28.68 & 18.6 & 14.73 & - \\
$DLBCL_A$ \cite{tissues} & 141 & 661 & 3 & 35.46 & 34.75 & 29.79 & - & - \\
$DLBCL_B$ \cite{tissues} & 180 & 661 & 3 & 48.33 & 28.33 & 23.33 & - & - \\
$TCGA$ \cite{weinstein2013cancer} & 801 & 20531 & 5 & 37.45 & 18.23 & 17.6 & 16.98 & 9.74 \\
    \bottomrule
\end{tabular}
\end{table}

\subsection{Experimental Evaluation}

To assess the performance of our proposed methodology, we adopted a stratified 10-fold cross-validation strategy. This approach ensures that each fold includes a representative distribution of the target classes, reducing potential bias in the evaluation process. The dataset was randomly partitioned into 10 subsets, each containing an approximately equal distribution of samples from every class. We performed training and testing of our model iteratively, with each fold acting as the testing fold while the remaining nine folds were used for training. This process was repeated for all the folds, resulting in a robust evaluation of our approach. 

\subsubsection{Evaluation of our method}
For each fold of the 10-fold cross-validation, we first generate a graph for each fold. Since the nodes vary with each fold (thus the input to the GCN), the resulting sparsified graph differs across every cross-validation iteration.

\subsubsection{Baselines} 

In the experiments, we consider various baseline classifiers, ranging from simple linear classifiers such as decision trees (DTs) \cite{breiman2017classification}, oblique predictive clustering trees (SpyCTs) \cite{stepinik2020oblique}, and support vector machines (SVMs) \cite{cortes1995support} to ensemble methods such as random forests (RFs) \cite{breiman2001random} and XGBoost (XGB) \cite{chen2016xgboost}. 

Next, we explain the methods used to leverage signals from the unlabeled data to aid the model in model training.
We use three well-established linear latent space projection methodologies, t-SNE \cite{maaten2008visualizing}, UMAP \cite{mcinnes2018umap}, and SVD \cite{wold1987principal}, to reduce high-dimensional data into lower-dimensional representations.  These methodologies convert the problem space from the original to a latent space where we can learn from labeled and unlabeled instances. After applying dimensionality reduction, the methods convert the high-dimensional data into lower-dimensional spaces. In this study, we exclude the comparison with autoencoder networks due to the scarcity of the data.

In each cross-validation step, we first learn the shared lower-dimensional space of the whole dataset, learn a classifier (e.g., a DT or an RF) only on the train folds, and apply it to the test fold.

\section{Results}
\label{sec:results}

\subsection{Experimental results}

We extensively evaluated our method compared to the base 5 linear learners and their corresponding combinations for each problem. Table \ref{tab: Performance evaluation.} presents the results.  

\begin{table}[H]
\caption{Performance evaluation. The values in bold represent the best-performing method per dataset. The \emph{avg.} column represents the average performance of a method across the dataset. Our method shows the overall best performance.}
        \label{tab: Performance evaluation.}
    \centering

    \resizebox{\linewidth}{!}{\begin{tabular}{llllllllll|c}
\toprule
Dataset &    $Breast_A$ &     $Breast_B$ &      $DLBCL_A$&      $DLBCL_B$ &      $DLBCL_C$ &      $DLBCL_D$ &      $Multi_A$ &      $Multi_B$ &         $TCGA$ &  \emph{Avg.}  \\
Method     &                  &                  &                  &                  &                  &                  &                  &                  &      &       \\
\hline

\emph{ours}       &    0.939$_{0.05 }$ &    0.845$_{0.22 }$ &    \textbf{0.98$_{0.043} $} &   \textbf{0.956$_{0.065} $} &   \textbf{0.88$_{0.169} $} &   0.777$_{0.156 }$ &   0.951$_{0.066 }$ &   0.867$_{0.221 }$ &   0.996$_{0.006 }$ &  \textbf{0.910} \\
\hline
DT       &     0.89$_{0.13 }$ &    0.665$_{0.19 }$ &   0.723$_{0.121 }$ &   0.744$_{0.117 }$ &    0.57$_{0.216 }$ &   0.572$_{0.121 }$ &    0.94$_{0.066 }$ &    0.75$_{0.247 }$ &   0.978$_{0.023 }$ &  0.759 \\
DT-svd   &   0.867$_{0.102 }$ &    0.86$_{0.128 }$ &   0.937$_{0.058 }$ &   0.878$_{0.054 }$ &   0.657$_{0.181 }$ &    0.635$_{0.12 }$ &   0.872$_{0.134 }$ &   0.658$_{0.219 }$ &   0.938$_{0.037 }$ &  0.811 \\
DT-tsne  &     0.5$_{0.136 }$ &   0.205$_{0.127 }$ &   0.503$_{0.148 }$ &   0.578$_{0.158 }$ &   0.203$_{0.124 }$ &    0.21$_{0.112 }$ &    0.41$_{0.144 }$ &   0.117$_{0.183 }$ &   0.966$_{0.021 }$ &  0.410\\
DT-umap  &   0.927$_{0.082 }$ &   0.645$_{0.205 }$ &    0.943$_{0.07 }$ &   0.928$_{0.075 }$ &   0.673$_{0.171 }$ &   0.558$_{0.228 }$ &   0.931$_{0.078 }$ &    0.492$_{0.27 }$ &   0.993$_{0.011 }$ &  0.78 \\
\midrule

RF        &    0.889$_{0.07 }$ &    0.75$_{0.163 }$ &   0.944$_{0.076 }$ &   0.928$_{0.061 }$ &   0.757$_{0.202 }$ &   0.746$_{0.113 }$ &     \textbf{0.98$_{0.04} $} &   0.833$_{0.224 }$ &   0.995$_{0.006  }$ &  0.869 \\
RF-svd   &   0.929$_{0.065 }$ &    0.82$_{0.189 }$ &    0.95$_{0.046 }$ &    0.911$_{0.09 }$ &   0.783$_{0.198 }$ &   0.752$_{0.132 }$ &   0.931$_{0.119 }$ &   0.733$_{0.238 }$ &    0.981$_{0.01 }$ &  0.866\\
RF-tsne   &   0.653$_{0.134 }$ &    0.41$_{0.202 }$ &    0.568$_{0.14 }$ &     0.7$_{0.141 }$ &    0.22$_{0.149 }$ &   0.411$_{0.119 }$ &   0.497$_{0.172 }$ &   0.092$_{0.142 }$ &   0.992$_{0.008 }$ &  0.505 \\
RF-umap   &   0.919$_{0.075 }$ &    0.77$_{0.155 }$ &   0.964$_{0.048 }$ &   0.911$_{0.087 }$ &   0.807$_{0.194 }$ &    0.66$_{0.212 }$ &   0.913$_{0.081 }$ &   0.558$_{0.224 }$ &   0.998$_{0.005 }$ &  0.833\\
\midrule

SVM       &    0.56$_{0.158 }$ &    0.73$_{0.142 }$ &   0.96$_{0.065 }$ &   0.878$_{0.108 }$ &   0.847$_{0.139 }$ &   0.605$_{0.158 }$ &    0.97$_{0.064 }$ &     \textbf{0.9$_{0.153}$ } &   \textbf{0.999$_{0.004} $ } &  0.827 \\
SVM-svd    &   0.919$_{0.087 }$ &    0.71$_{0.145 }$ &     0.9$_{0.107 }$ &     0.9$_{0.116 }$ &   0.767$_{0.264 }$ &   0.768$_{0.137 }$ &     0.98$_{0.06 }$ &    0.75$_{0.171 }$ &   0.994$_{0.012  }$ &  0.854\\
SVM-tsne   &   0.507$_{0.148 }$ &    0.395$_{0.21 }$ &   0.512$_{0.145 }$ &   0.633$_{0.125 }$ &   0.177$_{0.188 }$ &   0.386$_{0.112 }$ &    0.34$_{0.079 }$ &     0.033$_{0.1 }$ &   0.995$_{0.006 }$ &  0.442\\
SVM-umap   &   0.929$_{0.079 }$ &    0.69$_{0.114 }$ &   0.964$_{0.048 }$ &     0.9$_{0.102 }$ &    0.78$_{0.183 }$ &   0.684$_{0.213 }$ &   0.922$_{0.125 }$ &   0.642$_{0.183 }$ &   0.998$_{0.005 }$ &  0.834 \\
\midrule

SpyCT      &    0.939$_{0.05 }$ &   0.675$_{0.157 }$ &   0.951$_{0.063 }$ &   0.944$_{0.043 }$ &    0.57$_{0.074 }$ &   0.638$_{0.137 }$ &    0.96$_{0.066 }$ &   0.342$_{0.058 }$ &   0.808$_{0.081 }$ &  0.759 \\
SpyCT-svd &   \textbf{0.959$_{0.05} $} &    \textbf{0.88$_{0.133} $}&   0.971$_{0.057 }$ &   \textbf{0.956$_{0.054} $ } &    0.86$_{0.155 }$ &   \textbf{0.784$_{0.181} $} &     0.98$_{0.06 }$ &    0.708$_{0.18 }$ &   0.996$_{0.006 }$ &  0.899 \\
SpyCT-tsne  &   0.641$_{0.119 }$ &   0.245$_{0.196 }$ &    0.63$_{0.102 }$ &   0.672$_{0.123 }$ &    0.36$_{0.154 }$ &    0.41$_{0.082 }$ &   0.482$_{0.167 }$ &    0.158$_{0.16 }$ &   0.996$_{0.006 }$ &  0.511 \\
SpyCT-umap &   0.948$_{0.072 }$ &   0.745$_{0.208 }$ &   0.971$_{0.047 }$ &   0.878$_{0.096 }$ &    0.81$_{0.181 }$ &   0.644$_{0.182 }$ &   0.914$_{0.081 }$ &   0.817$_{0.213 }$ &   0.998$_{0.005 }$ &  0.858 \\
\hline

XGB       &       0.9$_{0.1 }$ &   0.695$_{0.239 }$ &    0.859$_{0.11 }$ &   0.844$_{0.096 }$ &   0.687$_{0.159 }$ &   0.761$_{0.166 }$ &    0.94$_{0.066 }$ &     0.8$_{0.221 }$ &   0.994$_{0.012 }$ &  0.831 \\
XGB-svd    &   0.898$_{0.063 }$ &   0.835$_{0.176 }$ &   0.958$_{0.047 }$ &   0.933$_{0.085 }$ &    0.74$_{0.174 }$ &   0.713$_{0.182 }$ &   0.911$_{0.145 }$ &   0.758$_{0.225 }$ &   0.981$_{0.015 }$ &  0.859\\
XGB-tsne    &   0.663$_{0.089 }$ &    0.14$_{0.156 }$ &   0.567$_{0.113 }$ &   0.683$_{0.129 }$ &   0.243$_{0.138 }$ &     0.303$_{0.1 }$ &   0.466$_{0.171 }$ &   0.158$_{0.219 }$ &   0.992$_{0.006 }$ &  0.469\\
XGB-umap   &   0.908$_{0.107 }$ &   0.665$_{0.228 }$ &   0.957$_{0.065 }$ &   0.867$_{0.114 }$ &   0.657$_{0.194 }$ &    0.65$_{0.198 }$ &   0.933$_{0.098 }$ &   0.575$_{0.265 }$ &   0.998$_{0.005 }$ &  0.801 \\

\midrule

\bottomrule
\end{tabular}
}

\end{table}

The results are presented in Table \ref{tab: Performance evaluation.}, demonstrate the competitive performance of our method, outperforming the simple baselines DT, RF, SpyCT, and XGB consistently while achieving comparable results to the semi-supervised methods (where we introduced the unlabeled data and performed dimensionality reduction). The inherent local-structure learning dynamics and random initialization of t-SNE \cite{kobak2021initialization} render it less fitting for direct semi-supervised space learning tasks, resulting in lower performance when combined with the baseline learners.  The base SVM method was superior to other methods on the $TCGA$ and $Multi_B$ datasets. Notably, our method exhibited superior performance on the $DLBCL$ $A$, $B$, and $C$ datasets and came within a 2\% margin of the performance on the TCGA dataset. However, our method faced challenges when applied to the $Breast$ datasets, characterized by limited data availability. Consequently, the performance of our method was suboptimal in this particular scenario. The semi-supervised methods resulted in a substantial performance boost for the simpler methods. This enhancement enabled the SpyCT method to outperform all other methods on the $Breast$ $A$ and $B$ datasets and the $DLBCL_B$ dataset.

\subsubsection{Statistical tests}
\begin{figure}[h]
    \centering
    \resizebox{\textwidth}{!}{\includegraphics{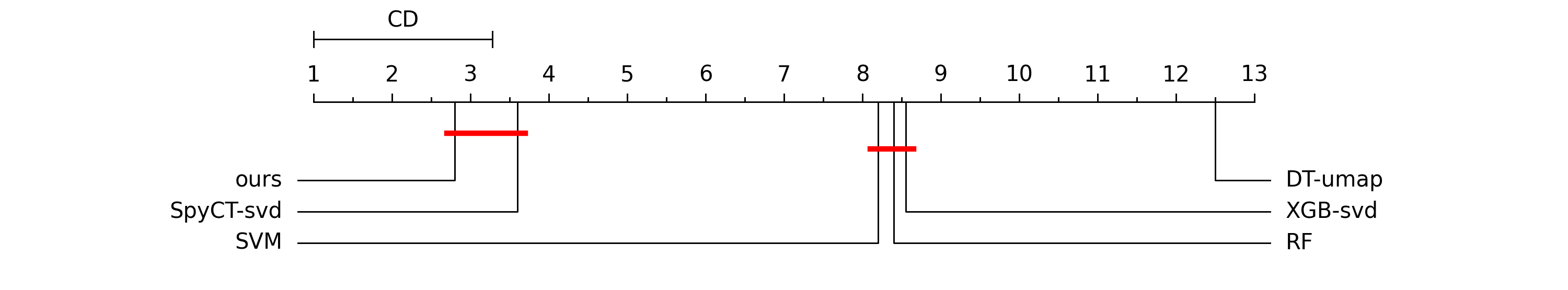}}
    \caption{Critical distance plot.}
    \label{fig:statistics-Nemenyi}
\end{figure}

We employ the Nemenyi test (Figure \ref{fig:statistics-Nemenyi}) with post-hoc correction \cite{demvsar2006statistical} at a significance level of 0.01. Red lines indicate statistically insignificant differences based on average scores. We choose the best-performing method for each model family, whether standalone or combined with a semi-supervised learner. Our method and the combination of SpyCT-SVD exhibit no statistically significant difference. The standalone SVM is the third-ranked method, which performs similarly to RF and XGB-SVD. Meanwhile, the Decision Tree, a more straightforward method, failed to beat the other models. More granular comparison can be seen on Figure \ref{fig:statistics-Nemenyi}.

We compared our method to the Random Forest and the SVM baseline using the \textit{Bayesian t-test} \cite{benavoli2017time}. We conducted 10 experiment runs for both methods, each testing the data on ten cross-validation folds. The probability of our model being better than the Random Forest was $90.57\%$, while the probability of both models being equal was $6.87\%$. By 'equal,' we mean they are within a $1\%$ margin of difference. As for the SVM model, it is better than ours with probability $0.55\%$, equivalent $0.7\%$, and worst with a probability of $98.73\%$

\begin{figure}
    \centering
    \begin{minipage}{0.5\textwidth}
        \centering
        \includegraphics[width=0.9\linewidth]{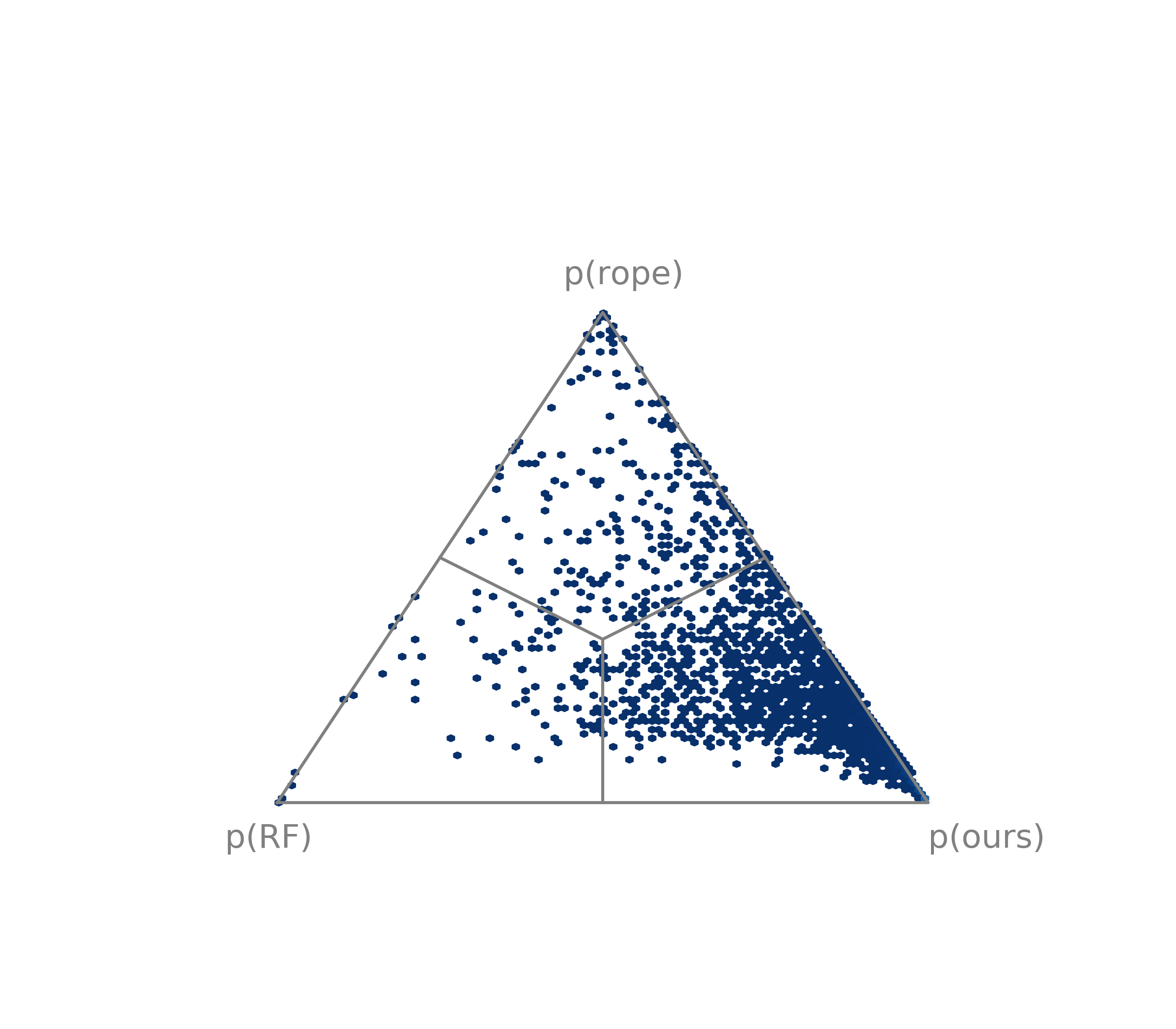}
        \caption{Our method and RF.}
        \label{fig:first_image}
    \end{minipage}\hfill
    \begin{minipage}{0.5\textwidth}
        \centering
        \includegraphics[width=0.9\linewidth]{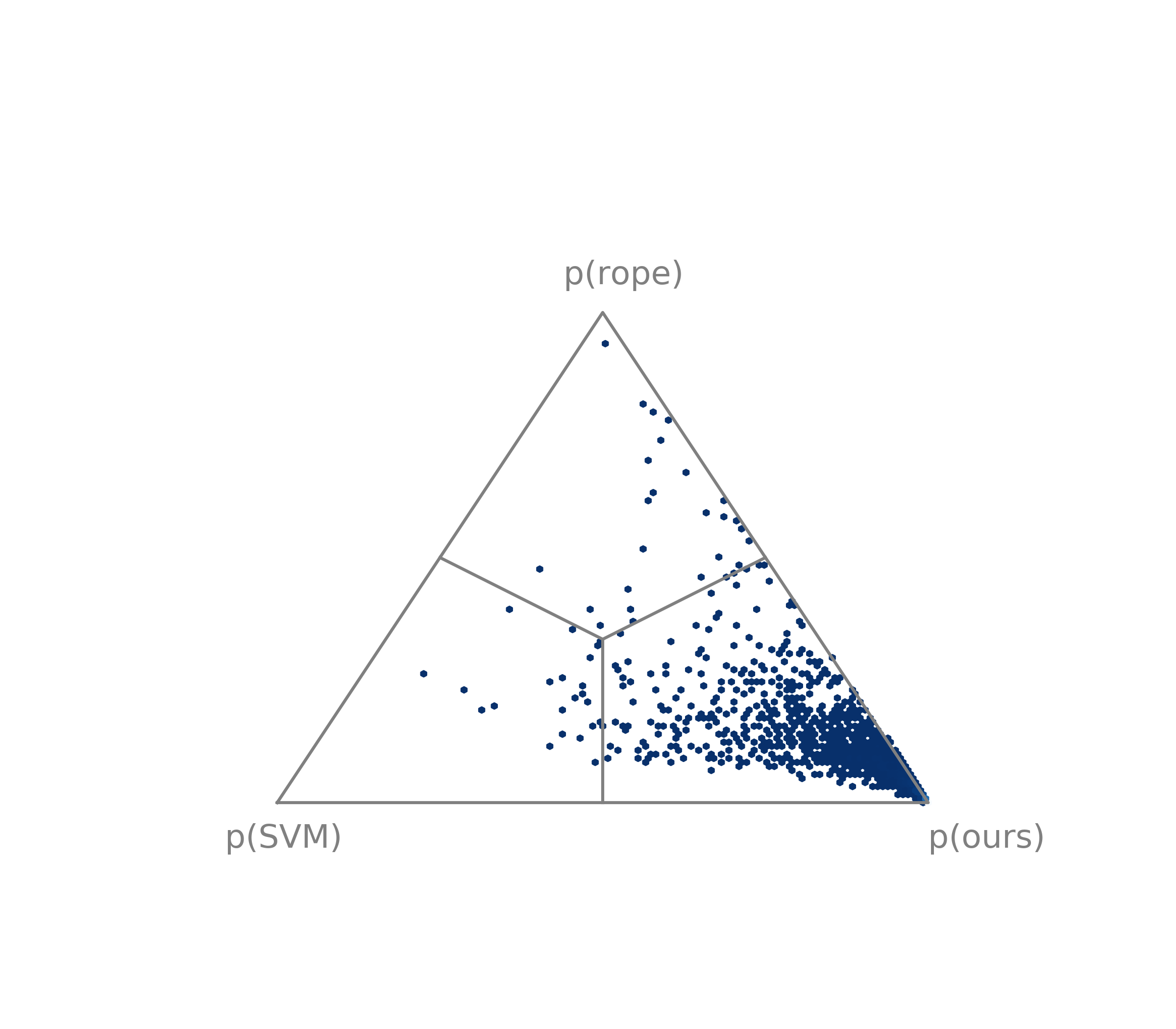} 
        \caption{Our method and SVM.}
        \label{fig:second_image}
    \end{minipage}
    \caption{Bayesian comparison of selected algorithm pairs.}
\end{figure}


\subsubsection{Time efficiency}

\begin{figure}[H]
        \centering
        \resizebox{0.5\textwidth}{!}{\includegraphics{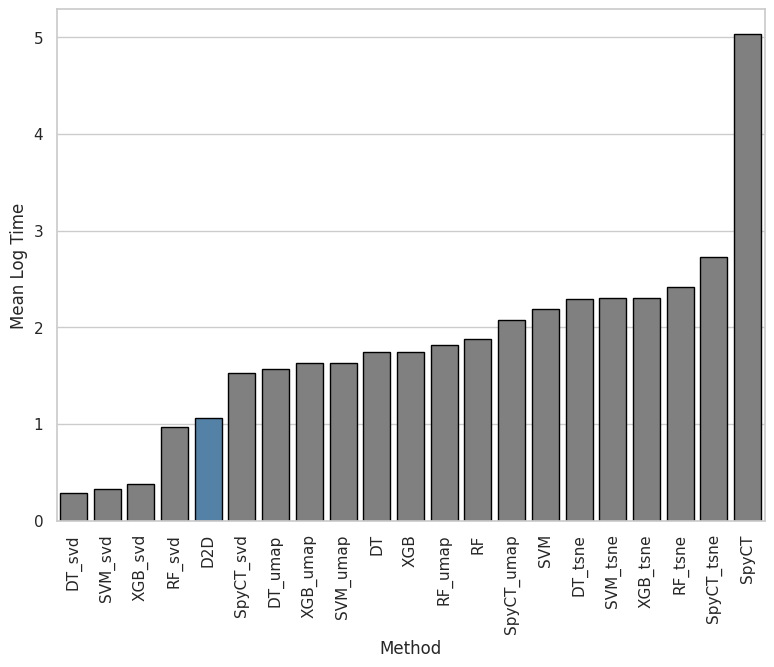}}
        \caption{Time comparison of the logarithm of the average time (in seconds) needed to learn a model, averaged across datasets. To enhance visibility, a logarithmic scale is applied. Our method is colored in blue, the lower the score the better.}
        \label{fig:time}
    \end{figure}

Next, we compared the time efficiency of our method to the baselines and the semi-supervised feature enrichment method. We measured the time for constructing the representations for each fold, learning on the training data, and predicting the test data. The results of the comparison are shown in Figure \ref{fig:time}. Our method outperformed all of the baselines time-wise on all of the features. Even when we applied lower space projection, our method still showed superior performance compared to other methods, except the application of SVD. 

\subsection{Abalation study}

\subsubsection{Latent Graph Structure}
In Table \ref{tab:net_stats}, we explore the latent graph structures determined by optimal thresholds for each dataset.
The $TCGA$ dataset, the largest one, has 801 nodes and 23,903 edges, showing that it's a dense network. This dataset has an assortativity of $0.74$, meaning that nodes with many connections tend to connect to other nodes with many connections. Also, its transitivity is $0.81$, showing a higher chance of triangle-shaped connections in the graph. Our approach achieved the most optimal scores on datasets $DLBCL_A$, $DLBCL_B$, and $DLBCL_C$. These datasets share notable transitivity, clustering coefficients, and closeness centrality. Suggesting local well-connected latent graphs where nodes are not only well-connected but can also access other nodes with minimal hops.
Only $DLBCL_{A}$ and $Multi_{A}$ were connected among all latent graphs. $DLBCL_{A}$ had a diameter of $3$ and an average shortest path of $1.71$, while $Multi_{A}$ had a diameter of $5$ and an average shortest path of $2.48$. However, datasets $DLBCL_D$ and $DLBCL_C$ stand out with lower clustering coefficients, hinting a reduced tendency towards cliquish behaviors.
Our method exhibits versatility as it performs effectively on connected and unconnected components.

\begin{table}[H]
    \centering
    \caption{Statistics of extracted latent graph structures, including metrics such as number of nodes, edges, homophily, heterophily, transitivity, assortativity, global clustering coefficient, average degree centrality, eigenvector centrality, closeness centrality, and number of connected components.}

   \resizebox{\textwidth}{!}{ \begin{tabular}{llrrrrrrrrrrr}
\toprule
   &  &  & &  & &  &  Global & Avg. & Avg. & Avg. & Num. \\
  Dataset & Nodes & Edges & Homophily & Heterophily & Transitivity & Assortativity & Clustering & Degree & Eigenvector & Closeness & Connected \\
   &  &  &  &   &  &  & Coefficient & Centrality & Centrality &  Centrality & Components \\
 
\midrule

$Multi_{B}$ & 32 & 14 &  0.36& 0.64 & 0.69 & -0.46 & 0.15 & 0.03 & 0.08 &  0.04 & 25 \\
$Breast_{B}$ & 49 & 63 &  0.70	& 0.30 &  0.52 & 0.49 & 0.35 & 0.05 & 0.07 &  0.08 & 15 \\
$DLBCL_{C}$ & 58 & 10 &0.90 & 0.10  & 0.27 & 0.18 & 0.04 & 0.01 & 0.04 &  0.01 & 49 \\
$Breast_{A}$ & 98 & 627 &0.93 & 0.07  &  0.68 & 0.45 & 0.52 & 0.13 & 0.06 &  0.30 & 5 \\
$Multi_{A}$ & 103 & 1002 & 0.84 & 0.16  &   0.73 & 0.48 & 0.69 & 0.19 & 0.05& 0.41 & 1 \\
$DLBCL_{D}$ & 129 & 41 & 0.59 &  0.41 &  0.40 & 0.30 & 0.05 & 0.00 & 0.02 & 0.01 & 100 \\
$DLBCL_{A}$ & 141 & 3075 & 0.72 &  0.28  & 0.57 & 0.06 & 0.57 & 0.31 & 0.08 & 0.58 & 1 \\
$DLBCL_{B}$ & 180 & 708 & 0.89 & 0.11 & 0.49 & 0.34 & 0.41 & 0.04 & 0.04 & 0.22 & 21 \\
$TCGA$  & 801 & 23903 & 0.98 & 0.02 & 0.80 & 0.74 & 0.66 & 0.07 & 0.01  & 0.31 & 10 \\
\bottomrule
\end{tabular}}
    \label{tab:net_stats}
\end{table}

\subsubsection{Latent Graph Similarity Thresholding}
To assess the thresholding parameter, $\theta$, we adopted a methodology based on selecting the parameter yielding the lowest training loss through early stopping.

\begin{figure} [H]
    \centering
    \resizebox{0.6\textwidth}{!}{\includegraphics{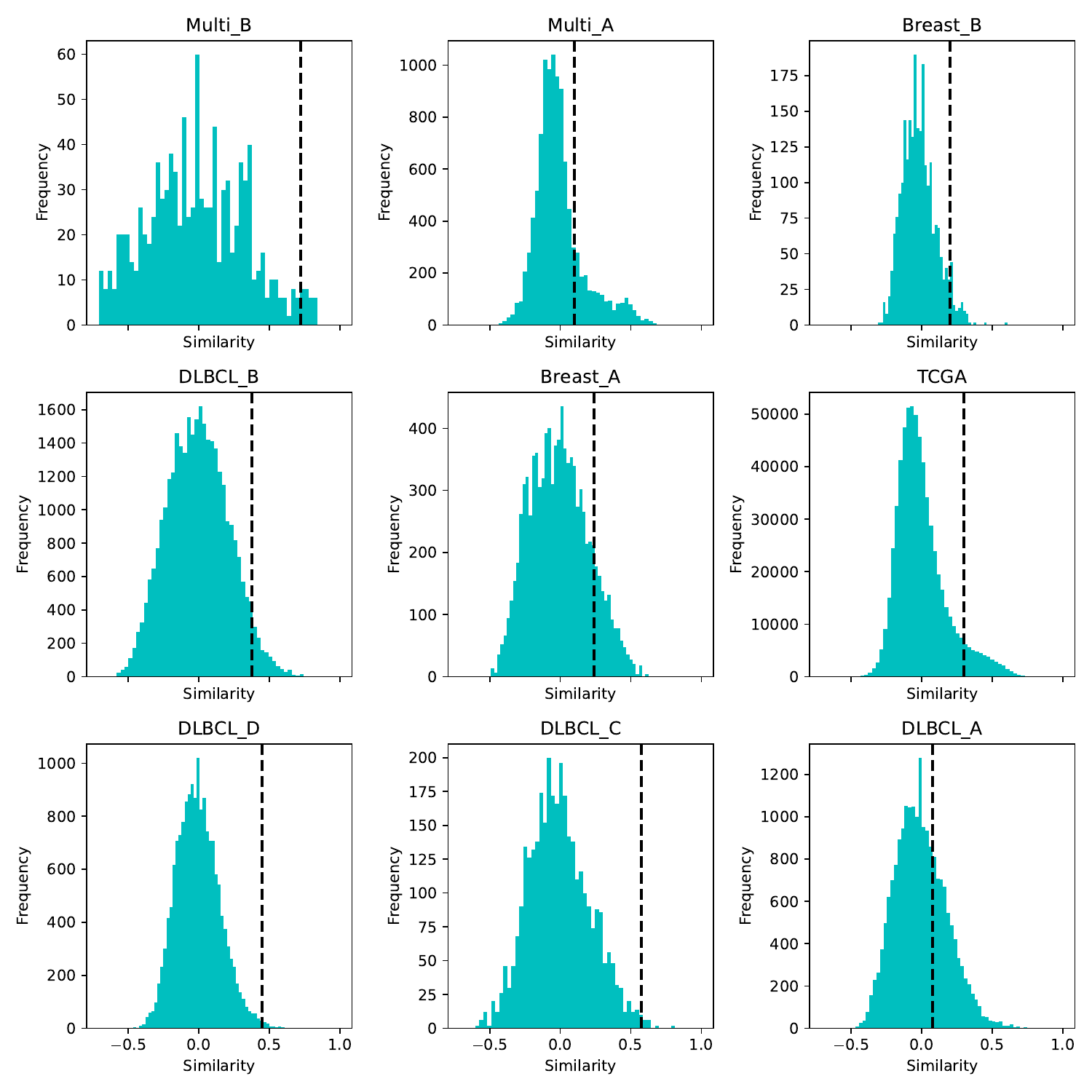}}
    \caption{The distribution of similarities and the threshold selected for constructing the latent graph based on cosine similarity.}
    \label{fig:enter-label}
\end{figure}

 For each dataset, we observed that different thresholding parameters appeared optimal. However, the optimal $\theta$ per dataset rendered graphs that contained 10\% to 15\% of the initially constructed edges in the full graph. Further insights regarding the thresholding parameter can be found in Figure \ref{fig:enter-label}.

\subsubsection{Qualitative evaluation}
\begin{figure}[H]
    \centering
    \resizebox{0.63\textwidth}{!}{\includegraphics{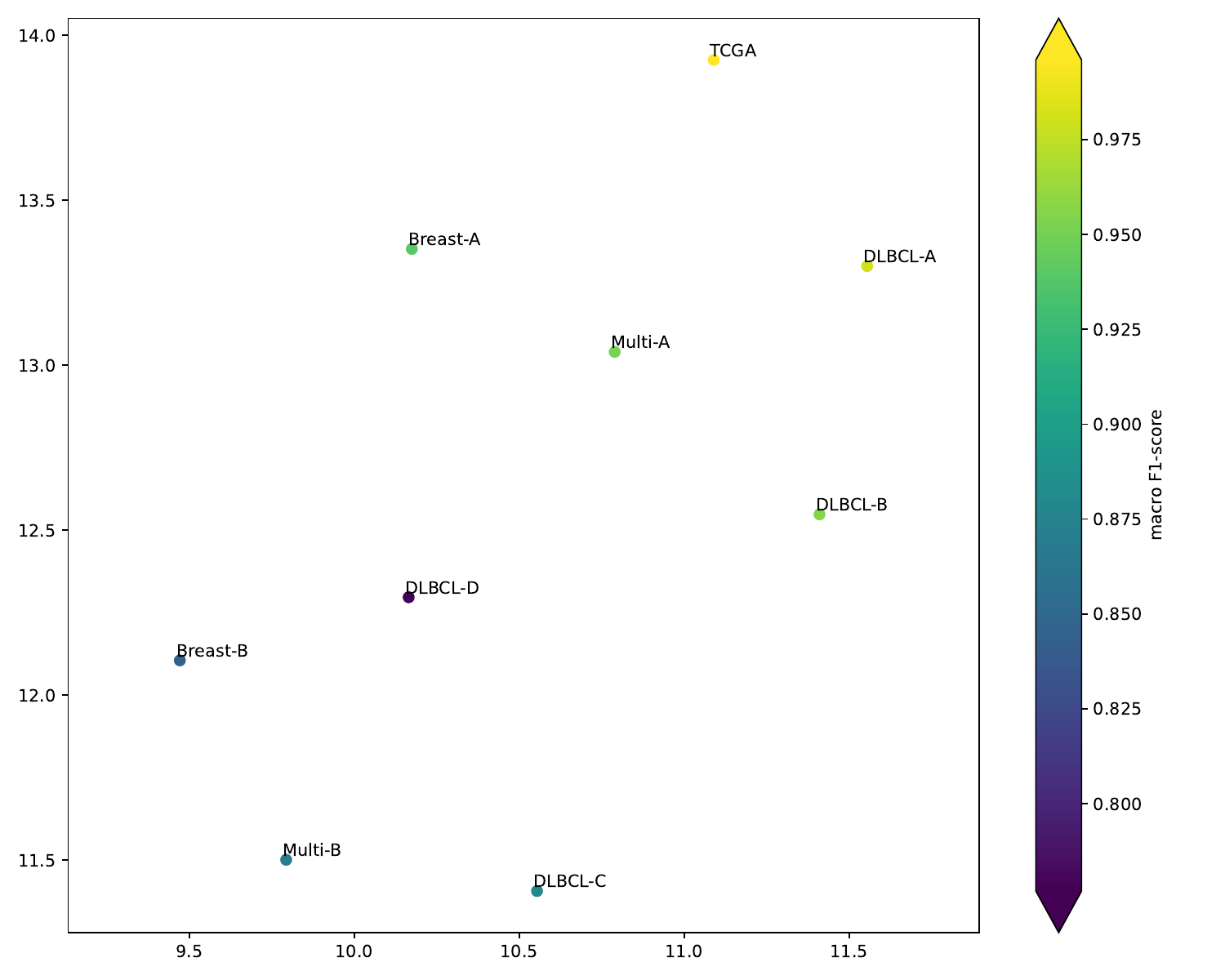}}
    \caption{Visualization of the embedded graphs of each dataset.}
    \label{fig:graphs_vec}
\end{figure}

 In Figure \ref{fig:graphs_vec}, the graphs are depicted based on their 2D low-dimensional projection and are color-coded by the macro F1-score. 

\begin{figure}[H]
    \centering
    \label{fig:graph_crt}
    \begin{minipage}{0.5\textwidth}
        \centering
        \includegraphics[width=0.9\linewidth]{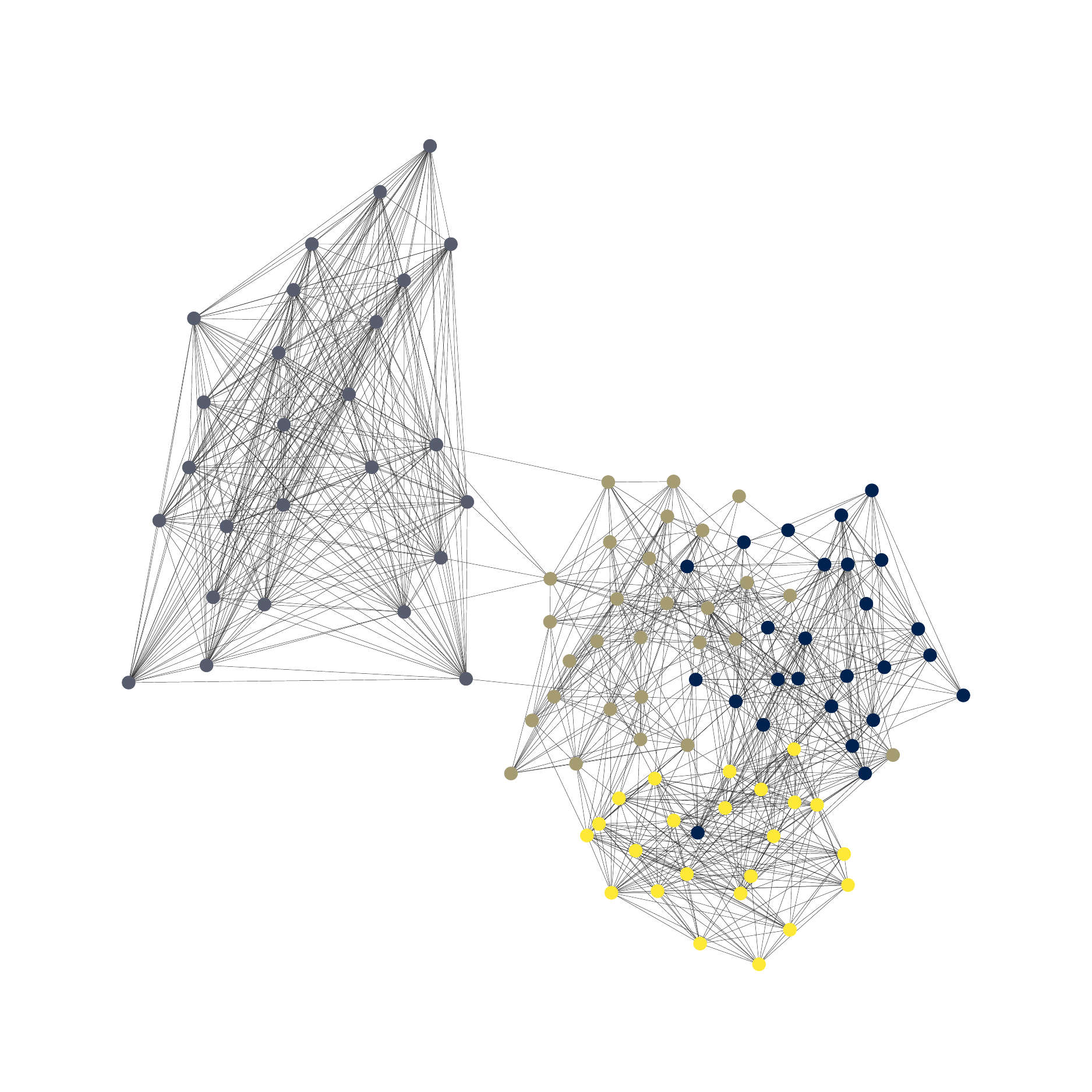}
        \caption{Visualization of Multi-A graph.}
        \label{fig:g1}
    \end{minipage}\hfill
    \begin{minipage}{0.5\textwidth}
        \centering
        \includegraphics[width=0.9\linewidth]{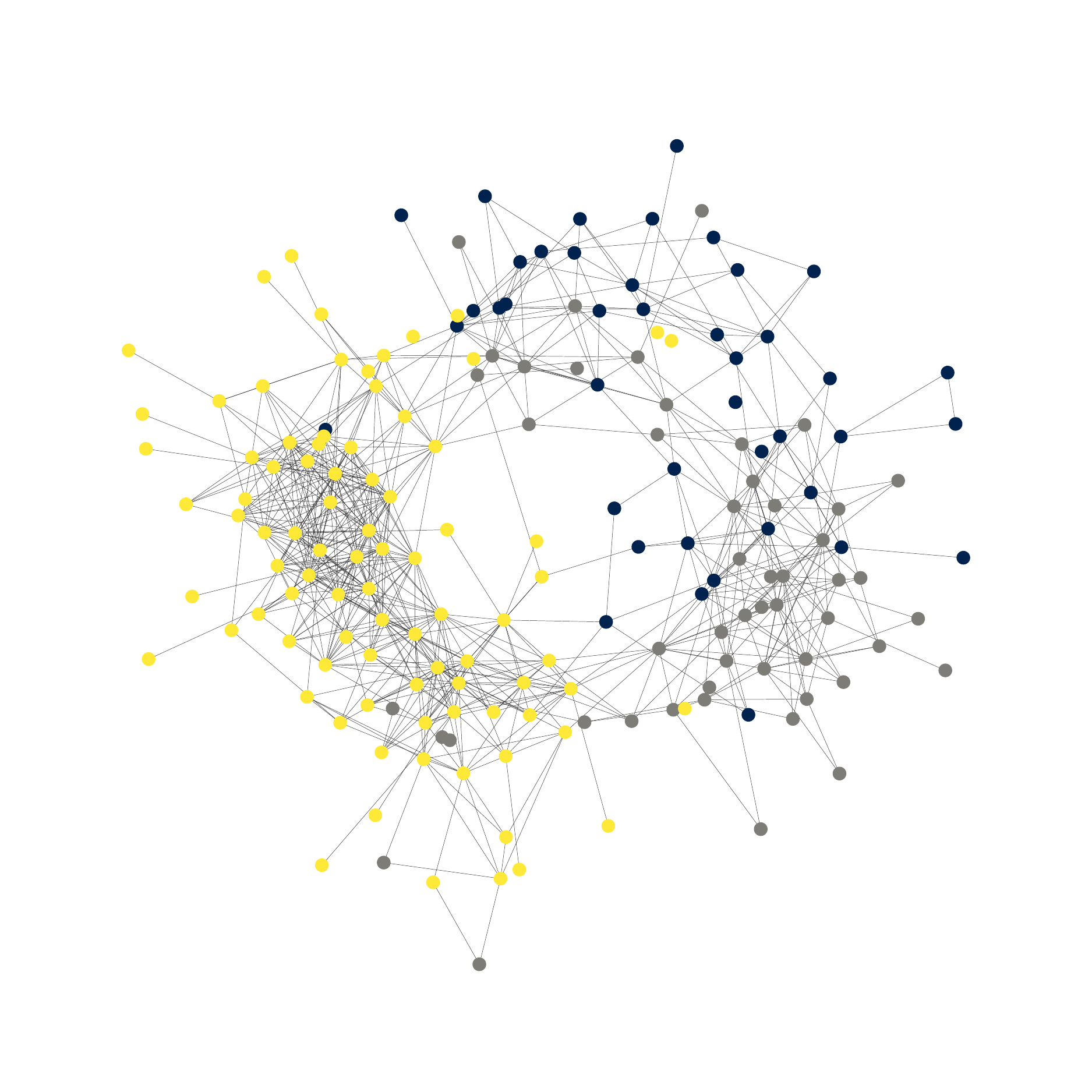} 
        \caption{Visualization of DLBCL-B graph.}
        \label{fig:second_image2}
    \end{minipage}
    \caption{Visualization of heterophilic and homophilic datasets represented as latent graphs.}
\end{figure}

Next, we explore semantic similarities between latent graphs. We embed the graphs using the Graph2Vec \cite{narayanan2017graph2vec} method with default parameters and subsequently embed them with UMAP \cite{mcinnes2018umap} in two dimensions. We observe that graphs with similar foundational structures cluster closely and perform similarly.

\section{Conclusions and Further Work}
In conclusion, our work presents a novel and time-efficient approach that leverages constructing a latent graph based on instance similarity and utilizes graph convolutional networks. The results obtained highlight the superior performance of our method in scenarios where instances are scarce and a large number of classes. Our method achieves this performance while maintaining computational efficiency, making it practical for real-world applications. For future work, we propose using physics-inspired methods to handle heterogeneity better. We also propose exploring approaches that perform automatic graph rewriting and thresholding, rather than the simple thresholding mechanism used in this work.

\section*{Acknowledgements}
he authors acknowledge the financial support from the Slovenian Research Agency through research core funding (No. P2-0103). Additionally, the first author's work was supported by a Young Researcher Grant PR-12394.
\section*{Availability}
The code and data to replicate the experiments are available on the following link \url{https://github.com/bkolosk1/latent_graph_tabular_data}.
%
%
\bibliographystyle{splncs04}
\bibliography{mybibliography}

\end{document}